\documentclass[journal]{IEEEtran}

\usepackage{cite}
\usepackage{graphicx}
\usepackage{color}
\usepackage{amsmath,bm}
\usepackage{url}
\usepackage{booktabs}

\def\I{{\cal I}}
\begin{document}

\title{Energy-based Dropout in Restricted Boltzmann Machines: Why not go random}

\author{Mateus Roder, \IEEEmembership{Member, IEEE}, Gustavo H. de Rosa, \IEEEmembership{Member, IEEE}, Victor Hugo C. de Albuquerque, \IEEEmembership{Senior Member, IEEE}, Andr\'e L. D. Rossi, \IEEEmembership{Member, IEEE}, and Jo\~{a}o P. Papa, \IEEEmembership{Senior Member, IEEE}
        
\thanks{Mateus Roder, Gustavo H. de Rosa, Andr\'e L. D. Rossi, and Jo\~{a}o P. Papa are with the S\~ao Paulo State University, Brazil and Victor Hugo C. de Albuquerque is with ARMTEC Tecnologia em Rob\'otica, Fortaleza/CE, Brazil. (email: \{mateus.roder, gustavo.rosa, andre.rossi, joao.papa\}@unesp.br, victor.albuquerque@ieee.org).}

\thanks{Manuscript received xx/xx/xxxx.}
}

\markboth{IEEE Transactions on Emerging Topics in Computational Intelligence}%
{Roder \MakeLowercase{\textit{et al.}}: Energy-based Dropout in Restricted Boltzmann Machines: Why not go random}

\maketitle

\begin{abstract}
Deep learning architectures have been widely fostered throughout the last years, being used in a wide range of applications, such as object recognition, image reconstruction, and signal processing. Nevertheless, such models suffer from a common problem known as overfitting, which limits the network from predicting unseen data effectively. Regularization approaches arise in an attempt to address such a shortcoming. Among them, one can refer to the well-known Dropout, which tackles the problem by randomly shutting down a set of neurons and their connections according to a certain probability. Therefore, this approach does not consider any additional knowledge to decide which units should be disconnected. In this paper, we propose an energy-based Dropout (E-Dropout) that makes conscious decisions whether a neuron should be dropped or not. Specifically, we design this regularization method by correlating neurons and the model's energy as an importance level for further applying it to energy-based models, such as Restricted Boltzmann Machines (RBMs). The experimental results over several benchmark datasets revealed the proposed approach's suitability compared to the traditional Dropout and the standard RBMs.
\end{abstract}

\begin{IEEEkeywords}
Machine learning, Restricted Boltzmann Machines, Regularization, Dropout, Energy-based Dropout
\end{IEEEkeywords}

\IEEEpeerreviewmaketitle

\section{Introduction}
\label{s.introduction}

Machine learning (ML) techniques have been broadly investigated to create authentic representations of the real world. Recently, deep learning has emerged as a significant area in ML~\cite{BengioTPAMI:13}, since its techniques have achieved outstanding results and established several hallmarks in a wide range of applications, such as image classification, object detection, and speech recognition, to cite a few.

Restricted Boltzmann Machines (RBMs)~\cite{Hinton:12} attracted considerable attention in the past years, mainly due to their simplicity, high-level parallelism, and comprehensive representation capacity. Such models stand for stochastic neural networks based on energy principles and guided by physical laws. Usually, these networks learn in an unsupervised fashion~\cite{Bengio:09} and are applied in various problems, e.g., image reconstruction, collaborative filtering, and feature extraction.

Machine learning algorithms are commonly trained according to an error metric called loss function (training error). Nevertheless, their biggest challenge lies in achieving a low generalization error (testing error). Whenever there is a high discrepancy between training and testing errors, the model expects to ``memorize" the training data, losing its generalization capacity and leading to reduced recognition rates when confronted with new data. One can acknowledge such a problem as overfitting.

Numerous attempts have been engaged in order to lessen the overfitting problem in classification tasks, such as early-stopping training or even introducing regularization methods such as soft-weight sharing~\cite{Nowlan:92}, L1~\cite{Tibshirani:96}, and L2~\cite{Hoerl:70}, DropConnect~\cite{wan2013}, among others. Alternatively, the best way to employ a regularization method would be to average the predictions of all possible parameter configurations, weighing the possibilities and checking out which would perform better. Nevertheless, such a methodology demands a cumbersome computational effort, only feasible for pitiful or non-complex models~\cite{Xiong:11}.

Some years ago, an regularization approach known as Dropout was proposed by Srivastava et al.~\cite{Srivastava:14} and aimed to turn off learning neurons using a random Bernoulli distribution. In other words, neurons and their outgoing and incoming connections are temporarily removed from the network according to a probability, allowing the evaluation of distinct sub-architectures and providing more robust training knowledge. Although it seems a straightforward method, the problem lies in that neurons are randomly dropped based only on a probability value ($p$), not taking advantage of valuable information related to the model itself. Also, the $p$ value have to be carefully chosen, since high probabilities of shutting off neurons may negatively impact the learning process.


Therefore, we aim to address such a problem through an energy-based Dropout, which creates a relationship between the system's neurons and its energy, removing standard Dropout's hyper-parameter ($p$) and the aleatory behavior while feeding in more robust information about the learning process itself.

In a nutshell, the main contributions of this paper are threefold: (i) to introduce a new type of regularization based on the model's energy, (ii) to introduce an energy-based Dropout in the context of RBMs, and (iii) to fill the lack of research regarding Dropout-based regularizations in RBMs. The remainder of this paper is organized as follows. Section~\ref{s.related_back} presents some studies and theoretical background concerning Dropout. Section~\ref{s.e_dropout} explains the energy-based Dropout, while Section~\ref{s.rbm} presents the central concepts of RBM, Dropout RBM, and energy-based Dropout RBM. Section~\ref{s.experimental_setup} discusses the experimental setup employed in this work, while Section~\ref{s.results} presents the experimental results. Finally, Section~\ref{s.conclusion} states conclusions and future works.
\section{Background and Related Work}
\label{s.related_back}

Dropout is a probability-based method~\cite{Srivastava:14} that decides whether a set of neurons should be dropped or not. This section presents the main concepts regarding such an approach and studies concerning such a regularization method.

\subsection{Related Works}
\label{ss.related}

Only a few recent studies have addressed the RBMs' overfitting problem with Dropout-based regularization. For instance, Wang et al.~\cite{Wang:13} have introduced a fast version of Dropout, but not aiming RBMs as their primary focus. The proposed approach is employed in classification and regression tasks and works by sampling from a Gaussian approximation instead of applying the Monte Carlo ``optimization''.

Ba et al.~\cite{Ba:13} proposed an adaptive Dropout for training deep neural networks, which is achieved by computing local expectations of binary dropout variables and by calculating derivatives using backpropagation and stochastic gradient descent. The experiments showed that the method achieved low misclassification rates in the MNIST and NORB datasets, highly competitive with CNNs.

Su et al.~\cite{Su:16} introduced a Dropout-based RBM considering field-programmable gate arrays, enabling improved implementation and hardware efficiency. Additionally, Wang et al.~\cite{Wang:17} presented an extensive review of different regularization methods in the context of RBMs, such as weight decay, network pruning, and Dropconnect. Although all these methods have obtained state-of-the-art results in some applications, their main drawback concerns setting up parameters.

Tomczak~\cite{Tomczak:16} employed different regularization methods for RBMs to improve their classification and generalization performance. In the experiments, the application of the considered regularization techniques did not result in any improvement. Nevertheless, when combining the information-theoretic regularization and the reconstruction cost, the proposed approach improved the log-probabilities.

In summary, RBM-related works show that when the main task is classification, such technique takes little advantages from the Dropout regularizer. On the other hand, it may boost the unsupervised learning, increasing the log-probabilities, and providing robustness data reconstruction. Considering that an RBM has a simple architecture, connections can quickly saturate, thus forcing the latent space to learn only the more prominent features from the data, which cause difficulties in data generalization and generation. It is interesting to employ an advanced regularization method, as the proposed approach, such that the energy associated with the latent representation indicates which hidden neurons need to be off to encourage others to learn more.

\subsection{Dropout Regularization}
\label{ss.dropout}

\begin{figure}[!ht]
    \centering
    \begin{tabular}{cc}
        \includegraphics[scale=0.5]{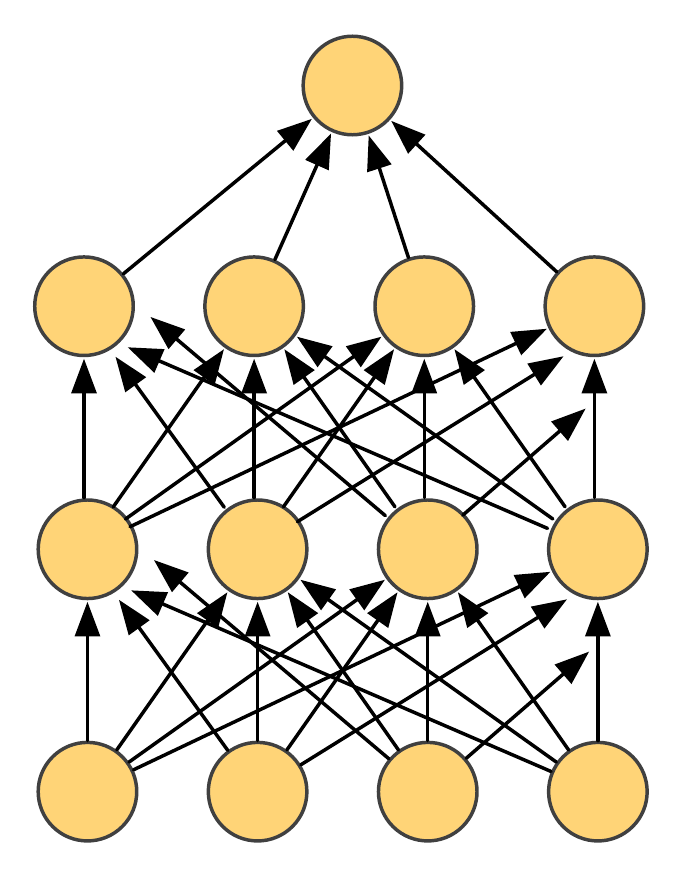} &
        \includegraphics[scale=0.5]{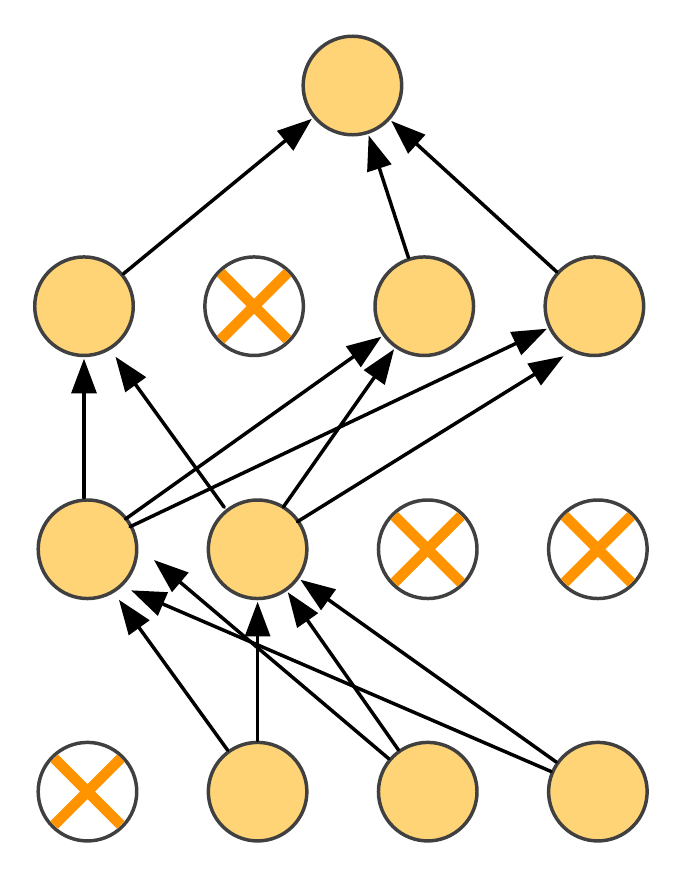} \\
        (a) & (b)    
    \end{tabular}
    \caption{Examples of: (a) standard network architecture and (b) a Dropout network architecture.}
    \label{f.networks}
\end{figure}

Dropout is a robust regularization method with a low computational cost that evaluates countless sub-architectures by randomly dropping out some neurons along the training process. Such a heuristic inhibits units from learning their neighbors' mistakes or ``memorizing'' the input data, been widely employed for classification tasks. Figure~\ref{f.networks} illustrates examples of both standard and Dropout network architectures.

Furthermore, it is straightforward to elucidate the mathematical foundations of Dropout. Let $\bm{r}$ be a vector of $n$ neurons of a specific layer $L$, where each variable $r_i$, $i = \{1, 2, \dots, n\}$, assumes the value 0 (zero) with probability $p$, regardless of other variables $r_j$, $j = \{1, 2, \dots, n\}$, where $i \neq j$. If $r_i=0$, the $i^{th}$ unit from the layer $L$ is temporarily switched-off alongside with its connections, while the unit is held when $r_i=1$.

Notice the probability $p$ is sampled directly from a Bernoulli distribution~\cite{Srivastava:14}, as follows:

\begin{equation}
\label{e.bernoulli}
r_i \sim Bernoulli(p), \forall i = \{1, 2, \dots, n\}.
\end{equation}
Besides, such a probability value is re-sampled for every batch during training.

Let $\gamma$ be the network activation function and $\bm{W}^{L}\in\Re^{m\times n}$ the weight matrix in a specific layer $L$. The activation vector $\bm{y}^{L}\in\Re^n$ can be formulated as follows:

\begin{equation}
\label{e.forward}
\bm{y}^{L} = \gamma(\bm{W}^{L}\bm{x}^{L}),
\end{equation}
where $\bm{x}^{L}\in\Re^m$ is the input from layer $L$.

In order to consider the dropout of neurons in this layer, the previous equation can be extended to the following:

\begin{equation}
\label{e.forward_dropout}
\bm{y}^{L} = \bm{r} \ast \gamma(\bm{W}^{L}\bm{x}^{L}),
\end{equation}
where $\ast$ stands for the point-wise operator.

Notably, the Dropout regularization provides training based on all possible $2^n$ sub-networks, as neurons are randomly shut down according to a probability $p$. Nevertheless, at the inference time (testing step), the weight matrix $\bm{W}^{L}$ needs to be re-scaled with $p$ in order to consider all possible sub-networks, as follows: 

\begin{equation}
\label{e.inference_dropout}
\bm{\tilde{W}}^{L} = p\bm{W}^{L}.
\end{equation}
\section{Energy-based Dropout}
\label{s.e_dropout}

In this section, we present the proposed approach denoted as energy-based Dropout (E-Dropout), which establishes a straightforward relationship between hidden neurons and the system's energy, hereinafter denoted ``Importance Level" ($\I$). The idea is to take advantage of the model's behavior for further enabling a more conscious decision whether a set of neurons should be dropped or not.

Let ${\cal{I}}^L\in\Re^n$ be the Importance Level of the hidden neurons at a specific layer $L$, which directly correlates the hidden probabilities with the RBM total energy. One can define ${\cal{I}}^L$ as follows:

\begin{equation}
\label{e.edrop}
{\cal{I}}^L = \dfrac{\bigg(\dfrac{P_{tr}(\bm{x}^L=1)}{P_{i}(\bm{x}^L=1)}\bigg)}{|\Delta E|},
\end{equation}
where $P_{tr}(\bm{x}^L=1)$ represents the probability of activating hidden neurons in layer $L$ after the training procedure, and $P_{i}(\bm{x}^L=1)$ stands for the activation probability of the hidden neurons in layer $L$ given the input data $\bm{x}$ only, i.e., before training. Finally, $|\Delta E|$ represents the absolute value of the system's energy variation, i.e., the energy after training subtracted from the initial energy measured.

The main intuition behind such a relationship derives from the RBM's energy, in which the hidden configuration participate directly to the total energy, as shown in Equation~\ref{e.energy_bbrbm} in the next section. The idea is to represent a gain or loss in information by applying a ratio between the pre- and post-neurons activation. Looking towards Equation~\ref{e.edrop}, one can observe an innovative way to model the relationship between neuron probability and the system's energy. In short, the meaning of a hidden neuron in the model is proportional to its importance level.

After computing $\cal{I}^L$ for each hidden neuron, it is possible to obtain the Dropout mask $\bm{s}$ by comparing it with a uniformly distributed random vector as follows:

\begin{equation}
\label{e.mask_edrop}
\bm{s} = \begin{cases}
1, & \text{if ${\cal{I}}^L < \bm{u}$} \\
0, & \text{otherwise,}
\end{cases}
\end{equation}
where $\bm{u}\in\Re^n$ is a uniformly distributed random vector, i.e., $\bm{u} \in [0,1)$. Furthermore, one can calculate the activation vector $\bm{y}^L$ as follows:

\begin{equation}
\label{e.mask_edrop2}
\bm{y}^{L} = \bm{s} \ast \gamma(\bm{W}^{L}\bm{x}^{L}).
\end{equation}

It is crucial to highlight that neurons tend to increase or decrease their importance level during the learning process based on the information acquired from the data distribution, where a neuron is less likely to be dropped out when its importance assumes a higher value. Additionally, when the system's energy is close to zero (more accurate data distribution learning), the energy-based Dropout allows a continuous drop out of neurons to learn additional information. Finally, during the inference phase, it is unnecessary to re-scale the weight matrix.
\section{Restricted Boltzmann Machines}
\label{s.rbm}

Restricted Boltzmann Machines~\cite{Hinton:02} are stochastic neural networks that deal with unlabeled data efficiently. In other words, RBMs are a suitable approach for unsupervised problems such as image reconstruction, feature extraction, pre-training deep networks, and collaborative filtering.

\begin{figure}[!ht]
    \centering
    \includegraphics[scale=0.65]{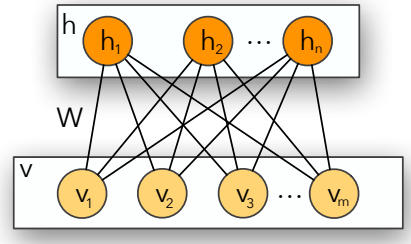}
    \caption{The standard RBM architecture.}
    \label{f.rbm}
\end{figure}

Such networks are modeled as bipartite graphs and parametrized by physical concepts like energy and entropy. Thereby, RBMs have a simple architecture with two binary-valued layers: the visible layer $\bm{v}$ with $m$ units, and the hidden layer $\bm{h}$ with $n$ units. Each connection between a visible $v_i$ and a hidden unit $h_j$  is weighted by $w_{ij}$. The weight matrix $\bm{W}_{m\times n}$ retains the knowledge of the network\footnote{Since RBMs have one hidden layer only, we omitted the layer index $L$.}. Figure~\ref{f.rbm} shows the standard architecture of an RBM.

While the visible layer handles the data, the hidden layer performs the feature extraction by detecting patterns and learning the data distribution in a probabilistic manner. Equation~\ref{e.energy_bbrbm} describes the energy function of an RBM, where $\bm{a}\in\Re^m$ and $\bm{b}\in\Re^n$ stand for the biases of visible and hidden units, respectively:

\begin{equation}
\label{e.energy_bbrbm}
E(\bm{v},\bm{h})=-\sum_{i=1}^ma_iv_i-\sum_{j=1}^nb_jh_j-\sum_{i=1}^m\sum_{j=1}^nv_ih_jw_{ij}.
\end{equation}

In addition, the joint probability of an arrangement $(\bm{v},\bm{h})$ can be modeled as folows:

\begin{equation}
\label{e.probability_configuration}
P(\bm{v},\bm{h})=\frac{e^{-E(\bm{v},\bm{h})}}{Z},
\end{equation}
where $Z$ is the partition function, which is a normalization term for the probability over all possible visible and hidden states. Moreover, the marginal probability of an input vector is represented as follows:

\begin{equation}
\label{e.probability_configuration2}
P(\bm{v})=\frac{\displaystyle\sum_{\bm{h}}e^{-E(\bm{v},\bm{h})}}{Z}.
\end{equation}

As in bipartite graph and in an undirected model, the activations for both units (visible and hidden) are mutually independent. Therefore, the formulation of their conditional probabilities is straightforward, being defined by as follows:

\begin{equation}
\label{e.pv}
 P(\bm{v}|\bm{h})=\prod_{i=1}^mP(v_i|\bm{h}),
\end{equation}
and
\begin{equation}
\label{e.ph}
    P(\bm{h}|\bm{v})=\prod_{j=1}^nP(h_j|\bm{v}),
\end{equation}
where $P(\bm{v}|\bm{h})$ and $P(\bm{h}|\bm{v})$ represent the probability of the visible layer given the hidden states and the probability of the hidden layer given the visible states, respectively.

From Equations~\ref{e.pv} and~\ref{e.ph}, we can derive the probability of a single active visible neuron $i$ given the hidden states, and the probability of a single active hidden neuron $j$ given the visible states, as follows:

\begin{equation}
\label{e.probv}
P(v_i=1|\bm{h})=\sigma\left(\sum_{j=1}^nw_{ij}h_j+a_i\right),
\end{equation}
and

\begin{equation}
\label{e.probh}
P(h_j=1|\bm{v})=\sigma\left(\sum_{i=1}^mw_{ij}v_i+b_j\right),
\end{equation}
where $\sigma(\cdot)$ stands for the logistic-sigmoid function.

Essentially, an RBM learns a set of parameters $\theta=(\bm{W}, \bm{a}, \bm{b})$ during the training process. Such task can be modeled as an optimization problem aiming to maximize the product of data probabilities for all training set ${\cal V}$, as follows:

\begin{equation}
\label{e.rbm_opt}
    \arg\max_{\Theta}\prod_{\bm{v}\in{\cal V}}P(\bm{v}).
\end{equation}

Such a problem is commonly treated by applying the negative of the logarithm function, known as the Negative Log-Likelihood (NLL), which represents the approximation of the reconstructed data regarding the original data distribution. Therefore, it is possible to take the partial derivatives of $\bm{W}$, $\bm{a}$ and $\bm{b}$ at iteration $t$. Equations~\ref{e.updateW},~\ref{e.updatea} and~\ref{e.updateb} describe the update rules for this set of parameters:

\begin{equation}
\label{e.updateW}
\bm{W}^{(t+1)} = \bm{W}^{(t)} + \eta(\bm{v}P(\bm{h}|\bm{{v}}) - \bm{\tilde{v}}P(\bm{\tilde{h}}|{\bm{\tilde{v}}})),
\end{equation}

\begin{equation}
\label{e.updatea}
\bm{a}^{(t+1)} = \bm{a}^{(t)} + (\bm{v} - \bm{\tilde{v}}),
\end{equation}
and
 
\begin{equation}
\label{e.updateb}
\bm{b}^{(t+1)} = \bm{b}^{(t)} + (P(\bm{h}|\bm{v}) - P(\bm{\tilde{h}}|\bm{\tilde{v}})),
\end{equation}
where $\eta$ is the learning rate, $\bm{\tilde{v}}$ stands for the reconstructed input data given $\bm{h}$, and $\bm{\tilde{h}}$ represents an estimation of the hidden vector $\bm{h}$ given $\bm{\tilde{v}}$. 

Hinton et al.~\cite{Hinton:02} proposed one of the most efficient ways to train an RBM and estimate the visible and hidden layers, known as the Contrastive Divergence (CD). Such an approach uses Gibbs sampling to infer the neurons' states, initializing the visible units with the training data.

\subsection{Dropout RBMs}
\label{ss.rbm_dropout}

\begin{figure}[!ht]
\centering
	\includegraphics[scale=0.65]{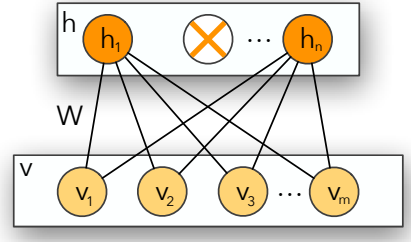}
	\caption{The Dropout-based RBM architecture.}
\label{f.rbm_dropout}
\end{figure}

Considering the concepts mentioned above, a Dropout RBM can be formulated as a simple RBM extended with one binary random vector $\bm{r} \in \{0, 1\}^n$. In this new formulation, $\bm{r}$ stands for the activation or dropout of the neurons in the hidden layer, where each variable $r_i$ determines whether the neuron $h_i$ is going to be dropped out or not. Figure~\ref{f.rbm_dropout} illustrates such an idea, in which the hidden unit $h_2$ is shutoff.

Notice that $\bm{r}$ is re-sampled for every mini-batch during learning. As units were dropped from the hidden layer, Equation~\ref{e.probh} can be rewritten as follows:

\begin{equation}
\label{e.probh_dropout}
P(h_j=1|\bm{r}, \bm{v})=
   \begin{cases}
      0, & \text{if}\ r_j=0 \\
      \sigma\left(\sum_{i=1}^mW_{ij}v_i+b_j\right), & \text{otherwise}.
    \end{cases}
\end{equation}

Therefore, a Dropout RBM can be understood as a blend of several RBMs, each one using different subsets of their hidden layers. As we are training the model with different subsets, the weight matrix $\bm{W}$ needs to be scaled at testing time, being multiplied by $p$ in order to adjust its weights (Equation~\ref{e.inference_dropout}).

\subsection{E-Dropout RBMs}
\label{ss.rbm_e-dropout}

As aforementioned in Section~\ref{s.e_dropout}, one can use Equation~\ref{e.edrop} to calculate the importance level $\cal{I}$ of the hidden neurons. Nevertheless, when dealing with an E-Dropout RBM, as the system's energy approximates to zero, $\cal{I}$ tends to overflow with large values. Therefore, it is necessary to re-scale $\cal{I}$ between $0$ and $1$ as follows:

\begin{equation}
\label{e.i_rescale}
{\cal{I}} = \dfrac{{\cal{I}}}{\max \{{\cal{I}}\}}.
\end{equation}

After computing $\cal{I}$, one can use Equation~\ref{e.mask_edrop} to calculate the Dropout mask $\bm{s}$. Therefore, Equation~\ref{e.probh} can be rewritten as follows:

\begin{equation}
\label{e.probh_edrop}
P(h_j=1|\bm{s}, \bm{v})=
   \begin{cases}
      0, & \text{if}\ s_j=0 \\
      \sigma\left(\sum_{i=1}^mW_{ij}v_i+b_j\right), & \text{otherwise}.
    \end{cases}
\end{equation}

Furthermore, it is worth using mini-batches while training the network, which can be accomplished by calculating Equation~\ref{e.i_rescale} for every sample in the mini-batch followed by its average.
\section{Experiments}
\label{s.experimental_setup}

In this section, we present the methodological setup used to evaluate the E-Dropout considering RBMs\footnote{RBMs, Dropout-RBMs, Weight-RBMs, and Energy-Dropout RBMs are available in Learnergy library~\cite{Roder:20}.} in the task of binary image reconstruction. Besides, we compare the proposed method against a standard-Dropout, RBMs without Dropout, among others, and describe the employed datasets and the experimental setup.

\subsection{Modeling E-Dropout RBMs}
\label{ss.modelling}

As aforementioned in Section~\ref{s.e_dropout}, the energy-based Dropout uses Equation~\ref{e.edrop} to calculate an importance level $\I$ for each neuron. Additionally, it computes the dropout mask $\bm{s}$ using Equation~\ref{e.mask_edrop}. Finally, it uses $\bm{s}$ in the same way as the standard Dropout method. Note that we consider the very same fundamental concepts presented in Section~\ref{s.rbm}.


\subsection{Datasets}
\label{ss.datasets}

Three well-known image datasets were employed throughout the experiments:

\begin{itemize}
\item MNIST\footnote{http://yann.lecun.com/exdb/mnist}~\cite{Lecun:98}: set of $28 \times 28$ grayscale images of handwritten digits (0-9), i.e., 10 classes. The original version contains a training set with $60,000$ images from digits `0'-`9', as well as a test set with $10,000$ images;

\item Fashion-MNIST\footnote{https://github.com/zalandoresearch/fashion-mnist}~\cite{Xiao:17}: set of $28 \times 28$ grayscale images of clothing objects. The original version contains a training set with $60,000$ images from $10$ distinct objects (t-shirt, trouser, pullover, dress, coat, sandal, shirt, sneaker, bag, and ankle boot), and a test set with $10,000$ images;

\item Kuzushiji-MNIST\footnote{https://github.com/rois-codh/kmnist}~\cite{Clanuwat:18}: set of $28 \times 28$ grayscale images of hiragana characters. The original version contains a training set with $60,000$ images from $10$ previously selected hiragana characters, and a test set with $10,000$ images.
\end{itemize}



\subsection{Experimental Setup}
\label{ss.setup}

Concerning the experimental setup, we employed five different RBM architectures, in which the main difference lies in the regularization method. In this case, RBM does not employ Dropout, the ``Weight"\ RBM (W-RBM) with L2 regularization employing a penalty of $5\cdot10^{-3}$ (the mean value of the ranges proposed by Hinton~\cite{Hinton:12}), the ``standard-dropout"\ RBM (D-RBM) uses the traditional Dropout, the ``DropConnect"\ RBM (DC-RBM) uses the traditional DropConnect, and the ``E-Dropout"\ RBM (E-RBM) employs the proposed energy-based Dropout. Additionally, when considering the ``standard-dropout" and the ``DropConnect", we used $p = 0.5$ as stated by Srivastava et al.~\cite{Srivastava:14} and Wan et al.~\cite{wan2013}, respectively.

Since the learning rate and the number of hidden neurons are important hyperparameters of an RBM, we fixed each RBM according to Table~\ref{t.rbm_parameters}, in which four different models have been considered, i.e., $M_a$, $M_b$, $M_c$, $M_d$. To provide more shreds of evidence of the E-Dropout suitability, we employed four distinct architectures, differing only in the number of hidden neurons and learning rates. 

Notice that three out of four architectures have $1,024$ hidden neurons. The reason is that RBMs with more feature detector units have more chances to learn unimportant information from the data distribution. Moreover, we decreased the learning rate to verify the E-Dropout ability to improve significantly when the network learns slowly.

Furthermore, we have considered $50$ epochs for the RBM learning procedure with mini-batches of size $256$, while all RBMs were trained using the Contrastive Divergence algorithm with $k=1$ (CD-1).

\begin{table}[!ht]
	\renewcommand{\arraystretch}{1.75}
	\caption{RBM hyperparameters configuration.}
	\label{t.rbm_parameters}
	\centering
	\begin{tabular}{lcccc}
	\toprule
	\textbf{Parameter} & $\mathbf{M_a}$ & $\mathbf{M_b}$ & $\mathbf{M_c}$ & $\mathbf{M_d}$\\
	\midrule
	$n$ (hidden neurons)   & 512 & 1,024 & 1,024 & 1,024 \\
	$\eta$ (learning rate) & 0.1 & 0.1 & 0.03 & 0.01     \\
	\bottomrule
	\end{tabular}
\end{table}

Two distinct metrics assessed the performance of the models on the test set, i.e., the Mean Squared Error (MSE) and the Structural Similarity Index (SSIM)~\cite{Wang:04}. The former is often known as the reconstruction error, which encodes the quality of the pixels reconstructed by the RBMs. In contrast, the latter provides a more efficient analysis of the image structure itself, which compares the quality between the original and reconstructed images.

To provide robust statistical analysis and acknowledge that the experiments' results are independent and continuous over a particular dependent variable (e.g., number of observations), we identified the Wilcoxon signed-rank test~\cite{Wilcoxon:45} satisfied our obligations. It is a non-parametric hypothesis test used to compare two or more related observations (in our case, repeated measurements of the MSE and SSIM values) to assess whether there are statistically significant differences between them or not. Therefore, we evaluated different RBM models with distinct Dropout methods ten times to mitigate the RBMs' stochastic nature for every dataset and architecture. Notice the statistical evaluation considers each model at once.

Finally, all the experiments were run in a desktop computer with 16 Gb of RAM ($2,400$MHz clock), an AMD processor containing six cores with 3 GHz of a clock, and a video card (GPU) GTX 1060 with 6 Gb of memory.
\section{Experimental Results}
\label{s.results}

This section presents the experimental results concerning the E-Dropout RBM, D-RBM, W-RBM, and RBM, considering three well-known literature datasets.

\subsection{MNIST}
\label{ss.mnist}

Considering the MNIST dataset, Table~\ref{t.mse_mnist} exhibits the mean reconstruction errors and their respective standard deviation over the testing set, where the best results are in bold according to the Wilcoxon signed-rank test. Considering model $M_b$ and $M_c$, the E-RBM could not obtain better results as the RBM, while for models $M_a$ and $M_d$, we can highlight that E-RBM was more accurate than RBM, W-RBM, D-RBM, and DC-RBM. Furthermore, these achievements evidence that E-Dropout is less sensitive to different learning rates.

\begin{table*}[!ht]
	\renewcommand{\arraystretch}{1.75}
	\caption{Mean reconstruction errors and their respective standard deviation on MNIST testing dataset.}
	\label{t.mse_mnist}
	\centering
	\begin{tabular}{lcccc}
		\toprule
		\textbf{Technique} & $\mathbf{M_a}$ & $\mathbf{M_b}$ & $\mathbf{M_c}$ & $\mathbf{M_d}$ \\ \midrule 
		RBM~\cite{Hinton:02} & 20.968 $\pm$ 0.059 & \textbf{17.728 $\pm$ 0.040} & \textbf{20.644 $\pm$ 0.039}  & 28.214 $\pm$ 0.068 \\
		W-RBM~\cite{Hinton:12} & 35.460 $\pm$ 0.126 & 31.753 $\pm$ 0.125 & 33.235 $\pm$ 0.062  & 39.583 $\pm$ 0.046 \\
		D-RBM~\cite{Srivastava:14} & 47.757 $\pm$ 0.242  & 47.968 $\pm$ 0.267 & 67.572 $\pm$ 0.0.339 & 80.155 $\pm$ 0.244 \\
		DC-RBM~\cite{wan2013} & 26.470 $\pm$ 0.134  & 25.595 $\pm$ 0.235 & 28.437 $\pm$ 0.142 & 35.856 $\pm$ 0.146 \\
		E-RBM & \textbf{20.511 $\pm$ 0.061} & 18.369 $\pm$ 0.072 & 21.277 $\pm$ 0.050  &\textbf{26.14 $\pm$ 0.174} \\
		\bottomrule
	\end{tabular}
\end{table*}

Table~\ref{t.ssim_mnist} exhibits the mean SSIM and their respective standard deviation over all experiments, where the best results are in bold according to the Wilcoxon signed-rank test. Considering model $M_b$ and $M_c$, the E-RBM obtained, statistically, the same results as the RBM, while for models $M_a$ and $M_d$, the E-RBM was significantly better than the RBM, W-RBM, D-RBM, and DC-RBM. It is interesting to note that the proposed approach overpass the other regularizers methods for all models, also for the architecture with $1,024$ hidden neurons and the lowest learning rate, the E-Dropout supported a $2.5\%$ performance improvement on SSIM, in front of an RBM (the second-best model).

\begin{table*}[!ht]
	\renewcommand{\arraystretch}{1.75}
	\caption{Mean SSIM values and their respective standard deviation on MNIST testing dataset.}
	\label{t.ssim_mnist}
	\centering
	\begin{tabular}{lcccc}
		\toprule
		\textbf{Technique} & $\mathbf{M_a}$ & $\mathbf{M_b}$ & $\mathbf{M_c}$ & $\mathbf{M_d}$ \\ \midrule
		RBM~\cite{Hinton:02} & 0.8170 $\pm$ 0.001            & \textbf{0.8410 $\pm$ 0.001}          & \textbf{0.8150 $\pm$ 0.000}          & 0.7490 $\pm$ 0.001        \\
		W-RBM~\cite{Hinton:12} & 0.7243 $\pm$ 0.001 & 0.7500 $\pm$ 0.001 & 0.7367 $\pm$ 0.001  & 0.6884 $\pm$ 0.001 \\
		D-RBM~\cite{Srivastava:14} & 0.5690 $\pm$ 0.002          & 0.5460 $\pm$ 0.003          & 0.3750 $\pm$ 0.002          & 0.2950 $\pm$ 0.001         \\
		DC-RBM~\cite{wan2013} & 0.7684 $\pm$ 0.001  & 0.7659 $\pm$ 0.002 & 0.7468 $\pm$ 0.001 & 0.6934 $\pm$ 0.001 \\
		E-RBM & \textbf{0.8240 $\pm$ 0.000} & \textbf{0.8410 $\pm$ 0.000}          & \textbf{0.8160 $\pm$ 0.001}          &\textbf{0.7740 $\pm$ 0.002}  \\
		\bottomrule
	\end{tabular}
\end{table*}

\begin{figure}[!ht]
\centering
\includegraphics[scale=0.5]{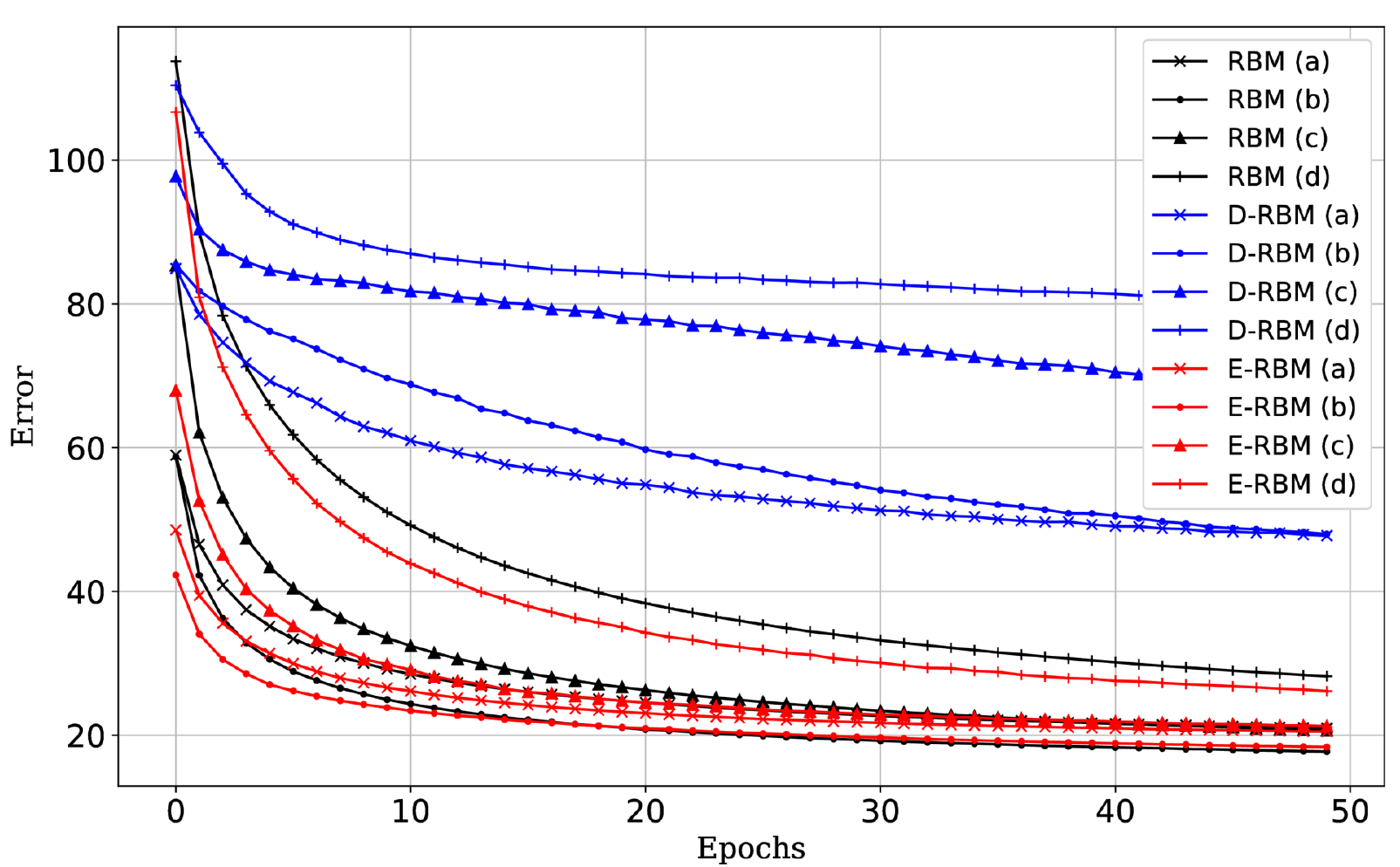}
\caption{Mean reconstruction error over the MNIST training set.}
\label{f.mnist_error}
\end{figure}

In summary, one can notice that E-RBM performed better than all the baselines with regularization. Additionally, Figure~\ref{f.mnist_error} depicts the mean reconstruction error over the training set only for models that employ Dropout regularization and the naive version (RBM) since such comparison stands for the work focus and more curves generate visually unattractive graphics. One can observe that most of the RBM models achieved better results than the D-RBM considering the same number of hidden neurons and learning rate. Nevertheless, for the models $M_a$ and $M_b$, the E-RBM achieved better reconstruction errors, besides converging faster at the first iterations.

\begin{figure}[!ht]
\centering
\includegraphics[scale=0.5]{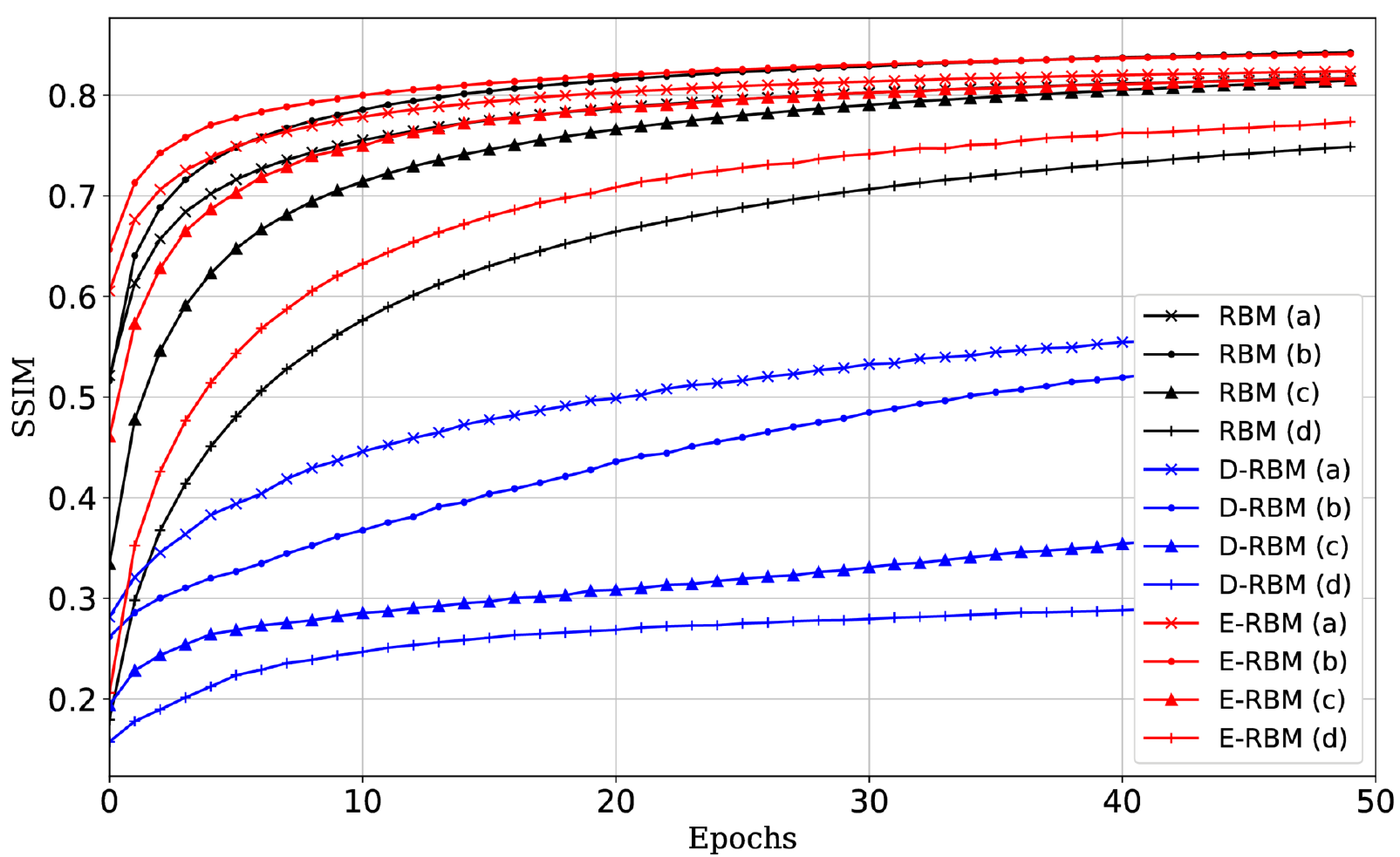}
\caption{Mean structural similarity index over the MNIST testing set.}
\label{f.mnist_ssim}
\end{figure}

Figure~\ref{f.mnist_ssim} depicts the mean SSIM over the testing set for both Dropout methods and the RBM naive version regarding all models. One crucial point to highlight is that all RBM and E-RBM models achieved better results than the D-RBM ones, probably due to the latter ``constant" neurons shutdown. Moreover, the E-Dropout achieved the best SSIM considering models $ M_a $, $M_b$, and $M_c$, thus fostering the proposed regularization technique.

\subsection{Fashion-MNIST}
\label{ss.fmnist}

Regarding the Fashion-MNIST dataset, Table~\ref{t.mse_fash} exhibits the mean reconstruction errors and their respective standard deviation over all experiments. Considering the E-RBM, it is clear its superiority regarding the baselines, once it achieved the lowest errors overall RBM architectures. We can highlight the performance on model $M_d$, which was $5.77\%$ better than standard RBM.

\begin{table*}[!ht]
	\renewcommand{\arraystretch}{1.75}
	\caption{Mean reconstruction errors and their respective standard deviation on Fashion-MNIST testing set.}
	\label{t.mse_fash}
	\centering
	\begin{tabular}{lcccc}
		\toprule
		\textbf{Technique} & $\mathbf{M_a}$ & $\mathbf{M_b}$ & $\mathbf{M_c}$ & $\mathbf{M_d}$ \\ \midrule
		RBM~\cite{Hinton:02} & 55.258 $\pm$ 0.097 & 53.204 $\pm$ 0.075 & 58.077 $\pm$ 0.066 & 67.293 $\pm$ 0.070 \\
		W-RBM~\cite{Hinton:12} & 66.195 $\pm$ 0.238 & 59.660 $\pm$ 0.152 & 62.769 $\pm$ 0.115 & 73.732 $\pm$ 0.114 \\        
		D-RBM~\cite{Srivastava:14} & 127.76 $\pm$ 0.694 & 104.78 $\pm$ 0.722 & 118.52 $\pm$ 0.782 & 119.95 $\pm$ 0.497 \\
		DC-RBM~\cite{wan2013} & 71.161 $\pm$ 0.105  & 68.983 $\pm$ 0.146 & 73.101 $\pm$ 0.190 & 80.538 $\pm$ 0.205 \\
		E-RBM & \textbf{53.858 $\pm$ 0.180} &\textbf{52.288 $\pm$ 0.095} &\textbf{55.064 $\pm$ 0.091} &\textbf{61.52 $\pm$ 0.085} \\
		\bottomrule
	\end{tabular}
\end{table*}

Table~\ref{t.ssim_fash} exhibits the results concerning the SSIM measure, been the best ones according to the Wilcoxon's signed-rank test in bold. We can observe that the E-RBM achieved better results than other baselines for models $M_b$ and $M_c$ (similar to DC-RBM). Surprisingly, the DC-RBM achieved better results regarding models $M_a$, $M_c$, and $M_d$, which was unexpected since such behavior was not observed in Table~\ref{t.mse_fash}.

\begin{table*}[!ht]
	\renewcommand{\arraystretch}{1.75}
	\caption{Mean SSIM values and their respective standard deviation on Fashion-MNIST testing set.}
	\label{t.ssim_fash}
	\centering
	\begin{tabular}{lcccc}
		\toprule
		\textbf{Technique} & $\mathbf{M_a}$ & $\mathbf{M_b}$ & $\mathbf{M_c}$ & $\mathbf{M_d}$ \\ \midrule
		RBM~\cite{Hinton:02} & 0.5630 $\pm$ 0.001 & 0.5940 $\pm$ 0.000 & 0.5410 $\pm$ 0.000 & 0.4760 $\pm$ 0.000 \\
		W-RBM~\cite{Hinton:12} & 0.5295 $\pm$ 0.001 & 0.5608 $\pm$ 0.001 & 0.5463 $\pm$ 0.001 & 0.4913 $\pm$ 0.001 \\
		D-RBM~\cite{Srivastava:14} & 0.2010 $\pm$ 0.002 & 0.2570 $\pm$ 0.002 & 0.2220 $\pm$ 0.002 & 0.2130 $\pm$ 0.001 \\
		DC-RBM~\cite{wan2013} & \textbf{0.5791 $\pm$ 0.001}  & 0.5856 $\pm$ 0.001 & \textbf{0.5729 $\pm$ 0.001} & \textbf{0.5405 $\pm$ 0.001} \\
		E-RBM & 0.5760 $\pm$ 0.001 &\textbf{0.6130 $\pm$ 0.001}  &\textbf{0.5710 $\pm$ 0.001}  &0.5150 $\pm$ 0.001  \\
		\bottomrule
	\end{tabular}
\end{table*}

\begin{figure}[!ht]
\centering
\includegraphics[scale=0.5]{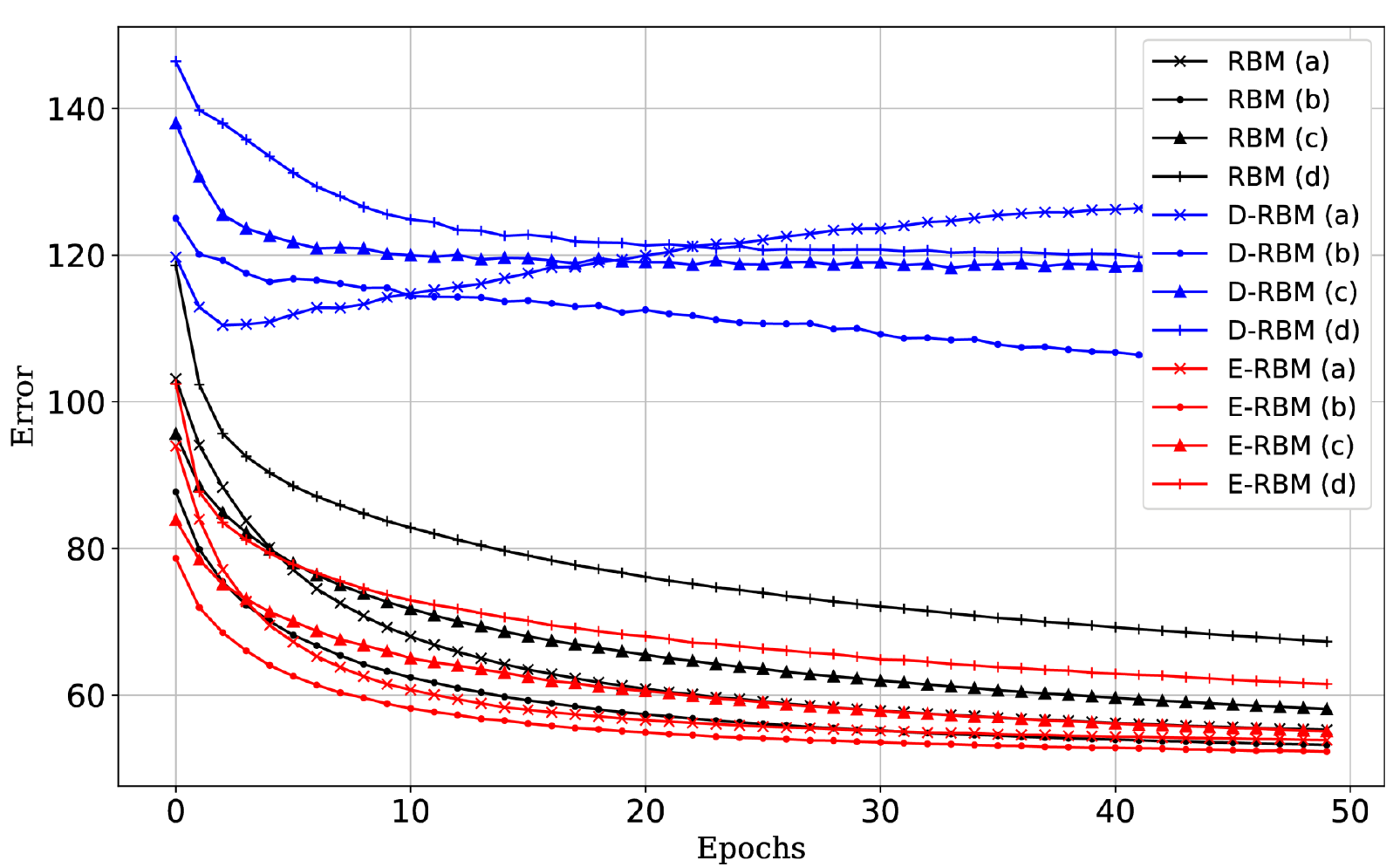}
\caption{Mean reconstruction error over Fashion-MNIST training set.}
\label{f.fash_error}
\end{figure}

Additionally, Figure~\ref{f.fash_error} depicts the mean reconstruction error for all architectures that employ Dropout and its naive version (RBM). One can note that the standard Dropout technique disturbed the RBM learning step. Moreover, the E-RBM achieved the best results in all models and the lowest reconstruction errors.

\begin{figure}[!ht]
\centering
\includegraphics[scale=0.5]{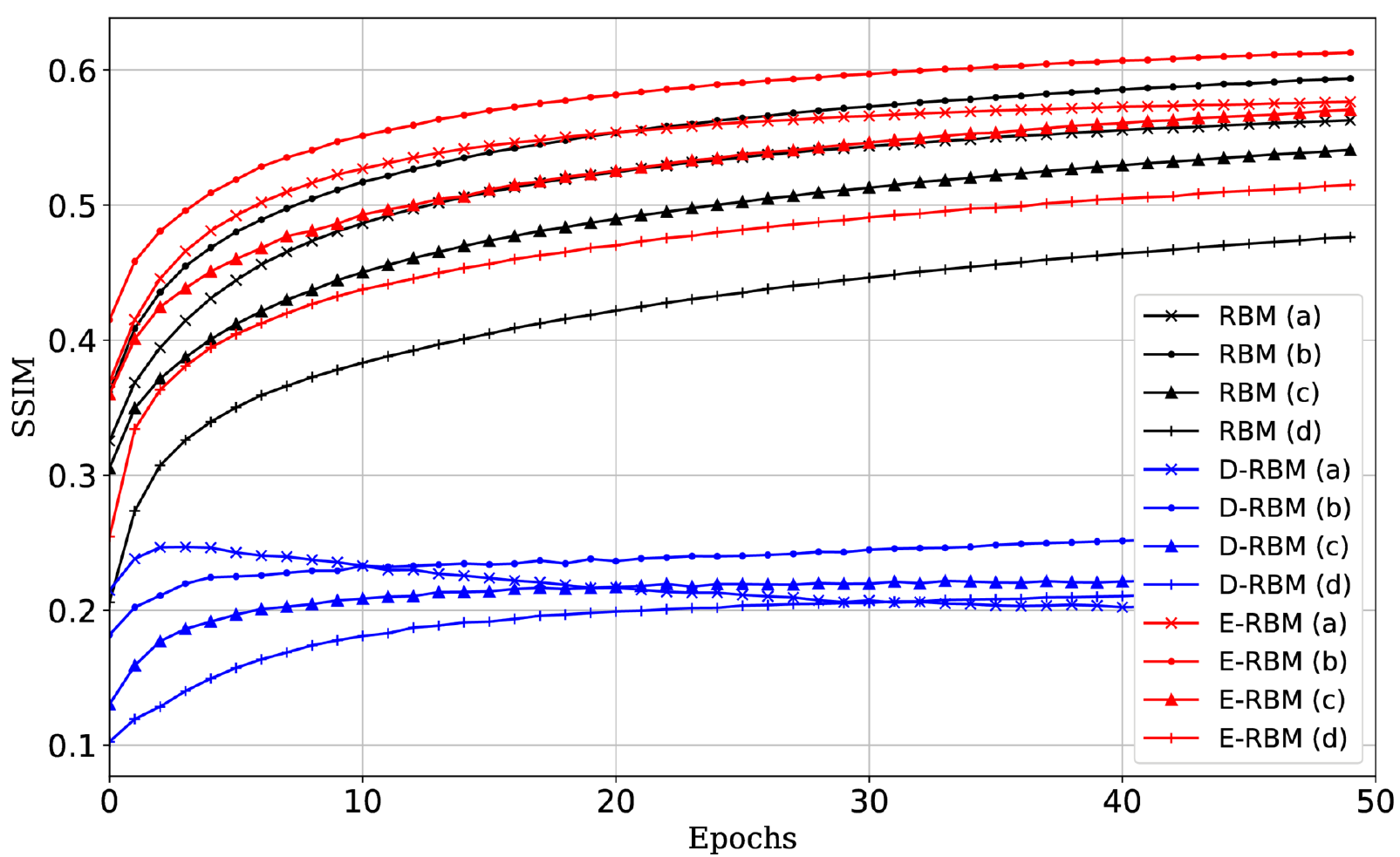}
\caption{Mean structural similarity index over Fashion-MNIST testing set.}
\label{f.fash_ssim}
\end{figure}

In the same manner, but assessing the models' performance by the SSIM measure (Figure~\ref{f.fash_ssim}), we can confirm that D-RBM was not able to be competitive against the energy-based Dropout and the RBM. Besides, E-RBM obtained better results than the RBMs without Dropout, strengthening its capability to increase the network's learning power.

\subsection{Kuzushiji-MNIST}
\label{ss.kmnist}

Considering the Kuzushiji-MNIST dataset, from Table~\ref{t.mse_kmnist}, one can see that the E-RBM achieved the lowest mean reconstruction errors for settings $M_a$ and $M_d$. For the model $M_b$, RBM was the best in front of the employed architectures, while for $M_c$, the difference between E-RBM and RBM was not significant. On the other hand, for model $M_d$, the E-RBM had a performance improvement of $3.32\%$ compared to RBM.

\begin{table*}[!ht]
	\renewcommand{\arraystretch}{1.75}
	\caption{Mean reconstruction error and their respective standard deviation on Kuzushiji-MNIST.}
	\label{t.mse_kmnist}
	\centering
	\begin{tabular}{lcccc}
		\toprule
		\textbf{Technique} & $\mathbf{M_a}$ & $\mathbf{M_b}$ & $\mathbf{M_c}$ & $\mathbf{M_d}$\\ \midrule
		RBM~\cite{Hinton:02} & 46.470 $\pm$ 0.121 & \textbf{37.587 $\pm$ 0.064} & \textbf{43.38 $\pm$ 0.070} & 58.262 $\pm$ 0.062 \\
		W-RBM~\cite{Hinton:12} & 76.839 $\pm$ 0.139 & 67.470 $\pm$ 0.139 & 70.537 $\pm$ 0.059 & 83.548 $\pm$ 0.061 \\        
		D-RBM~\cite{Srivastava:14} & 89.810 $\pm$ 0.249 & 80.475 $\pm$ 0.207 & 93.291 $\pm$ 0.189 & 109.436 $\pm$ 0.085 \\
		DC-RBM~\cite{wan2013} & 60.703 $\pm$ 0.149  & 53.330 $\pm$ 0.219 & 58.538 $\pm$ 0.141 & 74.056 $\pm$ 0.186 \\
		E-RBM & \textbf{44.853 $\pm$ 0.133} & 38.511 $\pm$ 0.086 & \textbf{43.544 $\pm$ 0.074} &\textbf{54.937 $\pm$ 0.143} \\
		\bottomrule
	\end{tabular}
\end{table*}

Additionally, Table~\ref{t.ssim_kmnist} exhibits the results for the SSIM metric. One can see that E-RBM kept the same previous behavior, meaning that for models $M_a$, and $M_d$, it achieved better results than the other baselines, while for model $M_b$ RBM and E-RBM have no statistical difference.

\begin{table*}[!ht]
	\renewcommand{\arraystretch}{1.75}
	\caption{Mean SSIM and their respective standard deviation on Kuzushiji-MNIST.}
	\label{t.ssim_kmnist}
	\centering
	\begin{tabular}{lcccc}
		\toprule
		\textbf{Technique} & $\mathbf{M_a}$ & $\mathbf{M_b}$ & $\mathbf{M_c}$ & $\mathbf{M_d}$\\ \midrule
		RBM~\cite{Hinton:02} & 0.6910 $\pm$ 0.001            & \textbf{0.7480 $\pm$ 0.001}         & 0.7040 $\pm$ 0.000          & 0.6060 $\pm$ 0.001        \\
		W-RBM~\cite{Hinton:12} & 0.5675 $\pm$ 0.001 & 0.6135 $\pm$ 0.001 & 0.5948 $\pm$ 0.001 & 0.5262 $\pm$ 0.001 \\
		D-RBM~\cite{Srivastava:14} & 0.3890 $\pm$ 0.002          & 0.4290 $\pm$ 0.002         & 0.2970 $\pm$ 0.002          & 0.2040 $\pm$ 0.000         \\
		DC-RBM~\cite{wan2013} & 0.6561 $\pm$ 0.001  & 0.6703 $\pm$ 0.001 & 0.6577 $\pm$ 0.001 & 0.5933 $\pm$ 0.001 \\
		E-RBM & \textbf{0.7040 $\pm$ 0.001} & \textbf{0.7480 $\pm$ 0.000}         &\textbf{0.7150 $\pm$ 0.000}  &\textbf{0.6370 $\pm$ 0.000}  \\
		\bottomrule
	\end{tabular}
\end{table*}

\begin{figure}[!ht]
\centering
\includegraphics[scale=0.5]{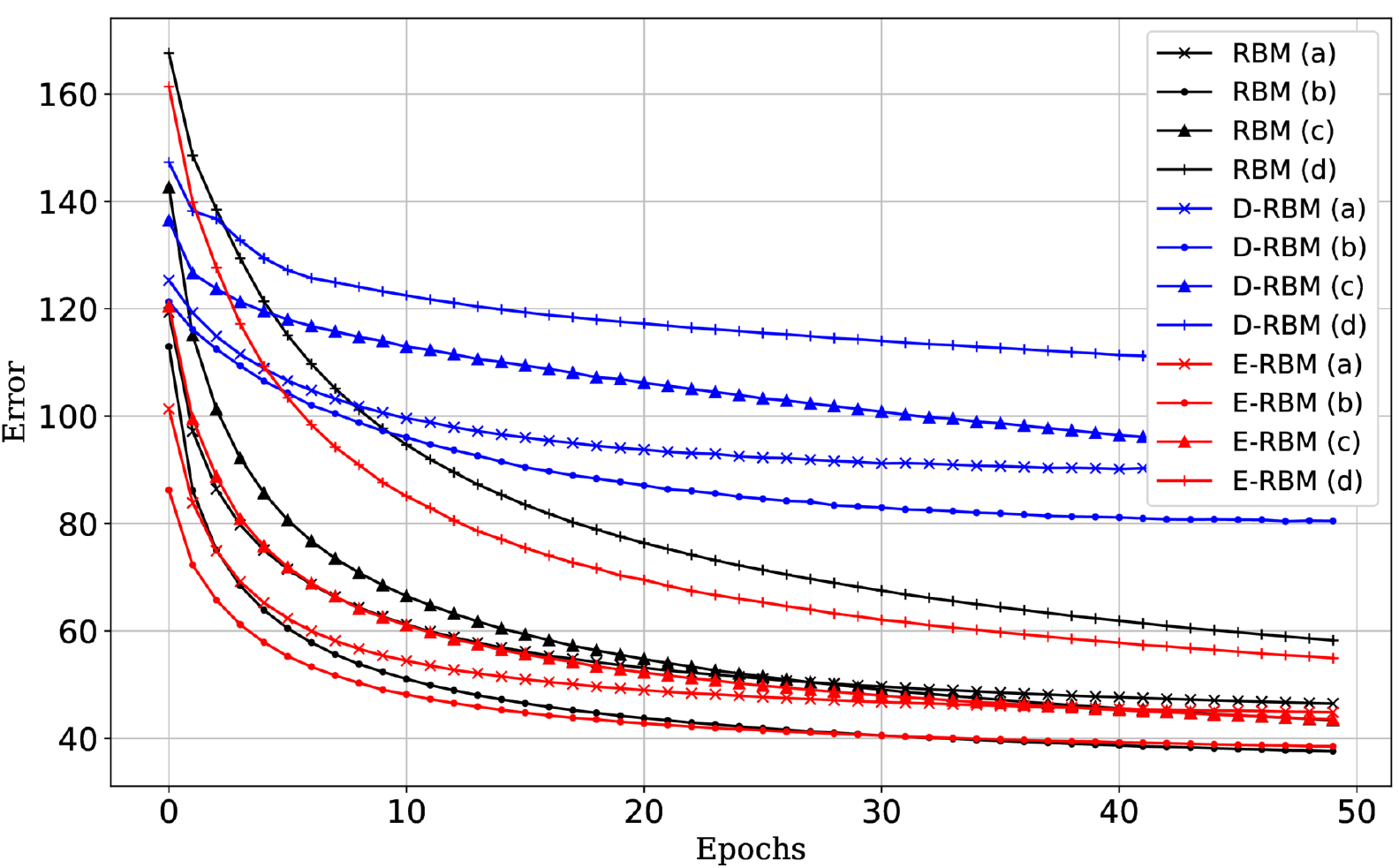}
\caption{Mean reconstruction error over Kuzushiji-MNIST's training set.}
\label{f.kmnist_error}
\end{figure}

Moreover, Figure~\ref{f.kmnist_error} depicts the mean reconstruction error over the training set for all the Dropout-based models and its naive version, respectively.
In this particular dataset, the MSE was considerably lower than Fashion-MNIST's ones, even though its digits seem more complex and have fewer details than Fashion-MNIST objects.

\begin{figure}[!ht]
\centering
\includegraphics[scale=0.5]{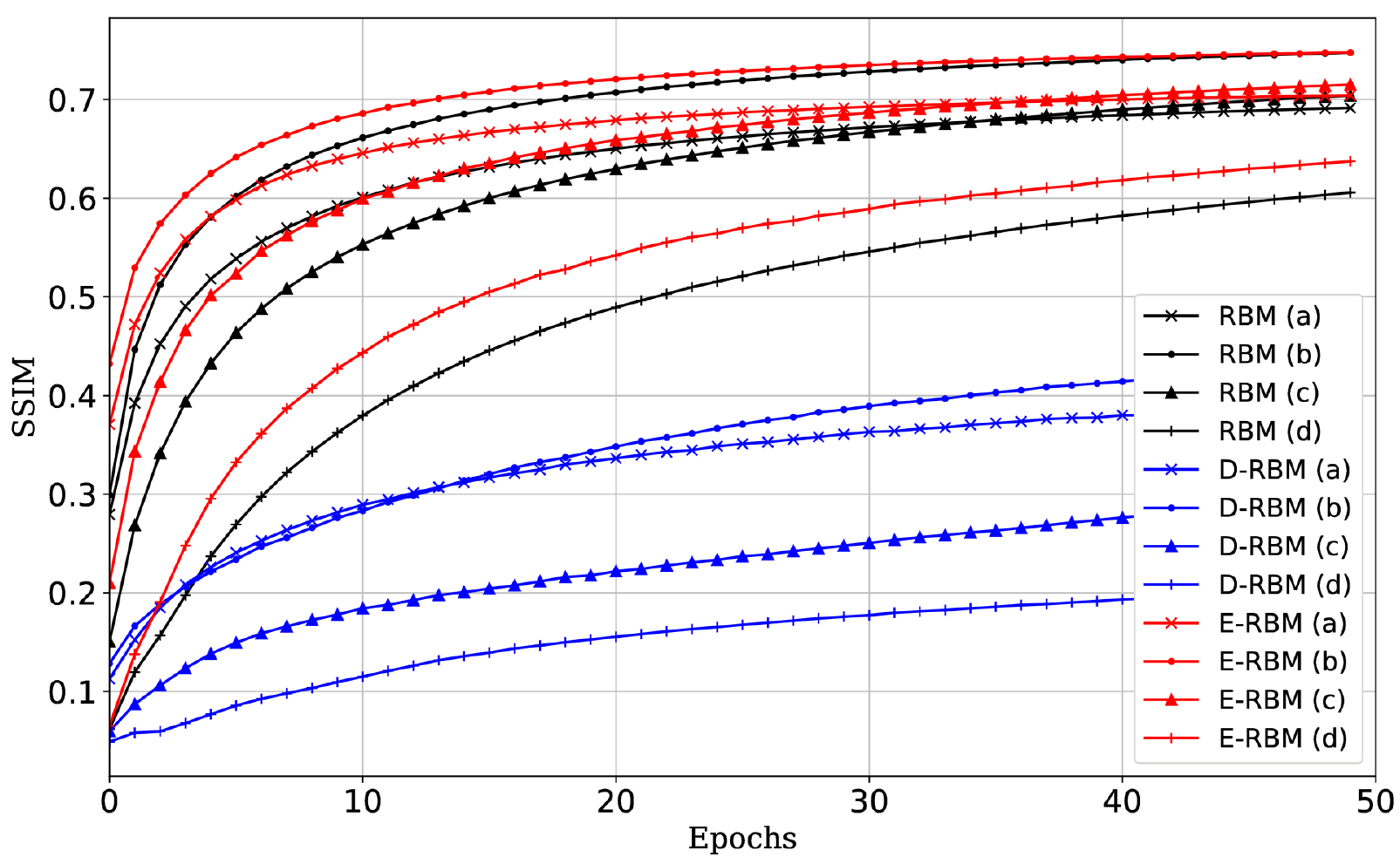}
\caption{Mean structural similarity index over Kuzushiji-MNIST's testing set.}
\label{f.kmnist_ssim}
\end{figure}

Furthermore, Figure~\ref{f.kmnist_ssim} exhibits the mean SSIM over the testing set considering the same approach that Figure~\ref{f.kmnist_error}. In this particular dataset, the E-RBM achieved almost the same performance as the RBM, for all configurations of $n$ (number of hidden neurons) and $\eta$ (learning rate).

\subsection{Dropout Overall Discussion}
\label{ss.discussion}

In addition to the reconstruction error and the visual quality of the reconstructed image assessed by SSIM, in this section, we analyze the behavior of E-Dropout related to the number of neurons dropped out over the training epochs and the weights learned by the E-RBM, D-RBM, and RBM models since it is the only ones employing neuron deactivation, in addition the original RBM. 

\begin{figure}[!ht]
\centering
\includegraphics[scale=0.5]{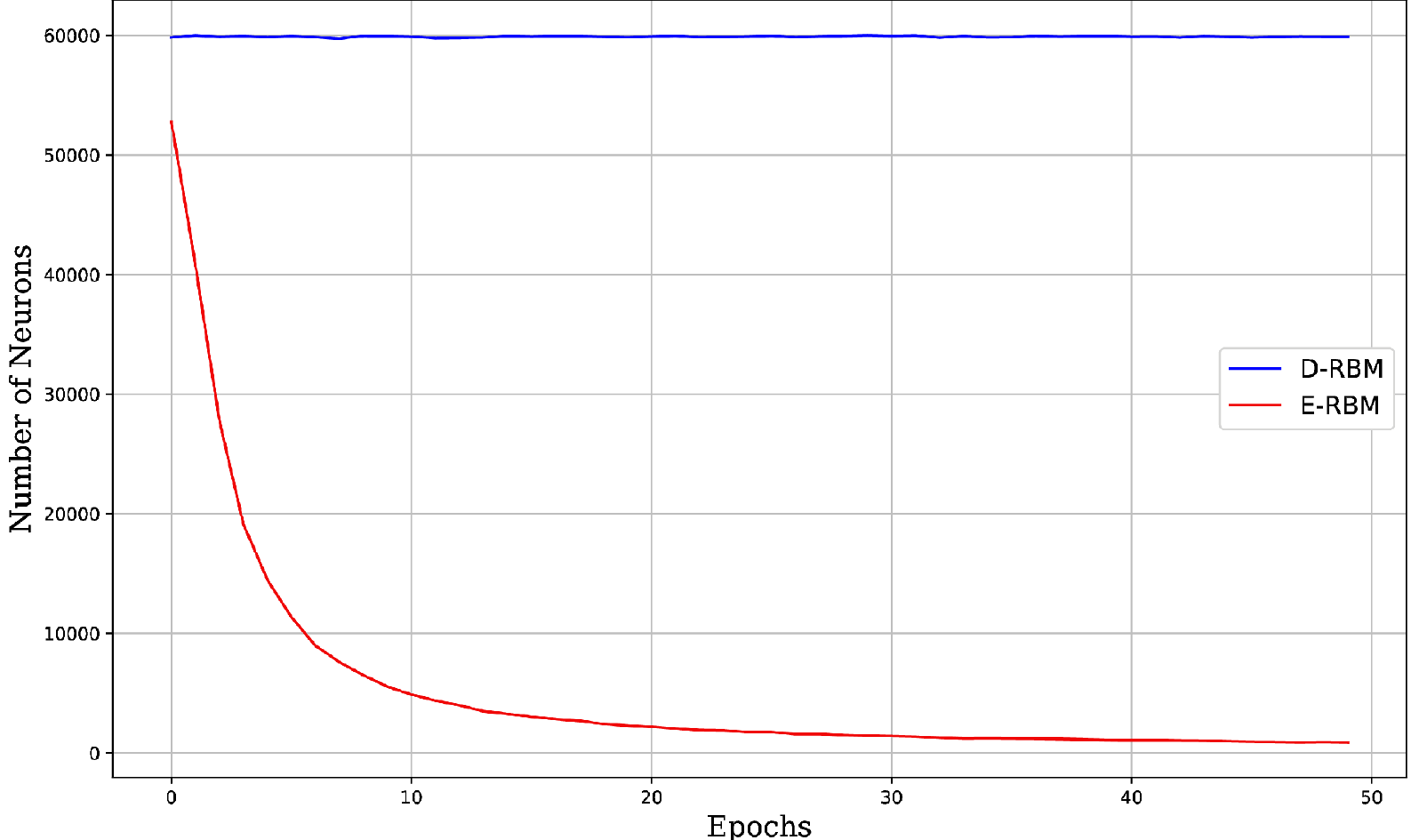}
\caption{Mean number of neurons dropped over MNIST's training set.}
\label{f.mnist_off}
\end{figure}
\begin{figure}[!ht]
\centering
\includegraphics[scale=0.5]{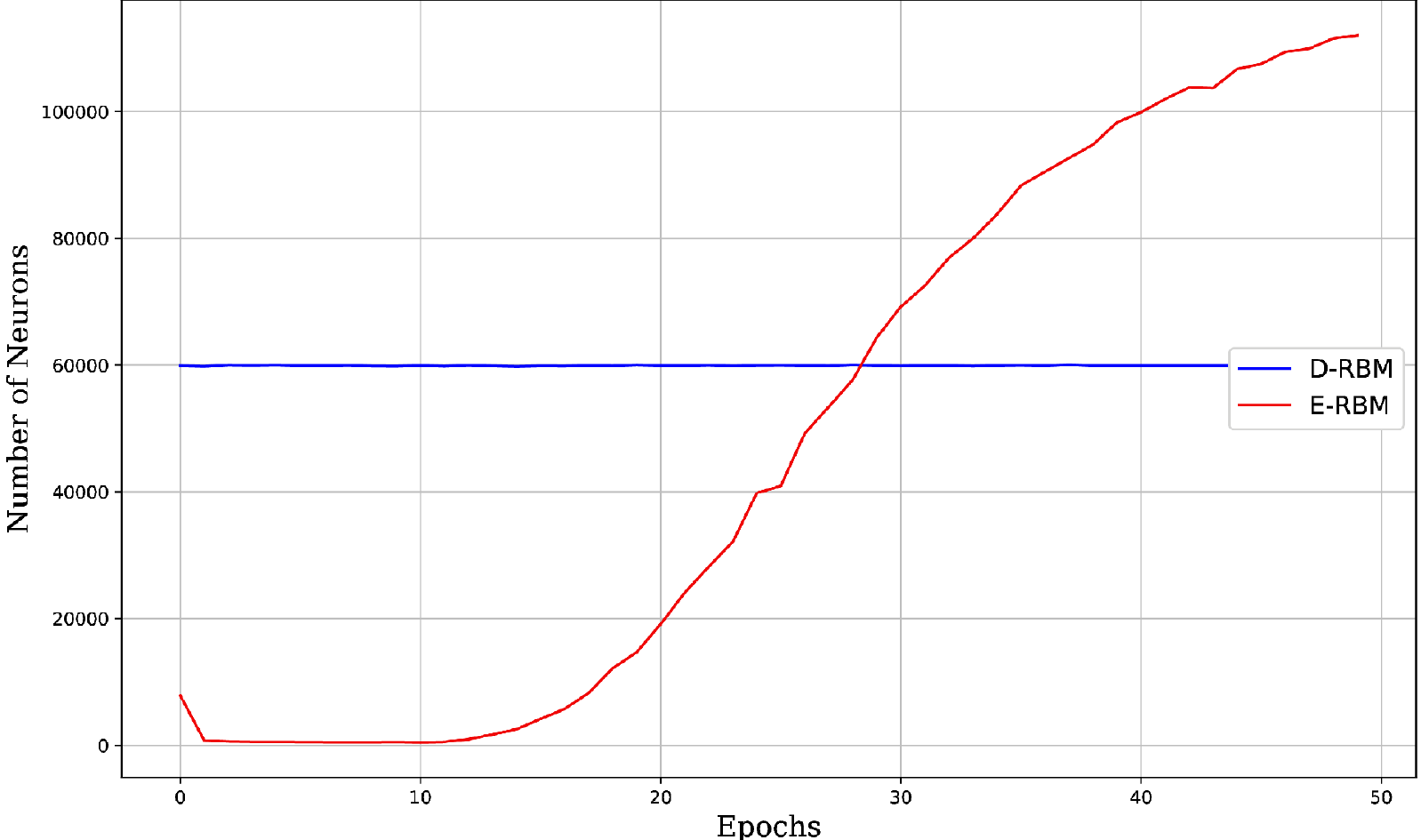}
\caption{Mean number of neurons dropped over Fashion-MNIST's training set.}
\label{f.fash_off}
\end{figure}
\begin{figure}[!ht]
\centering
\includegraphics[scale=0.5]{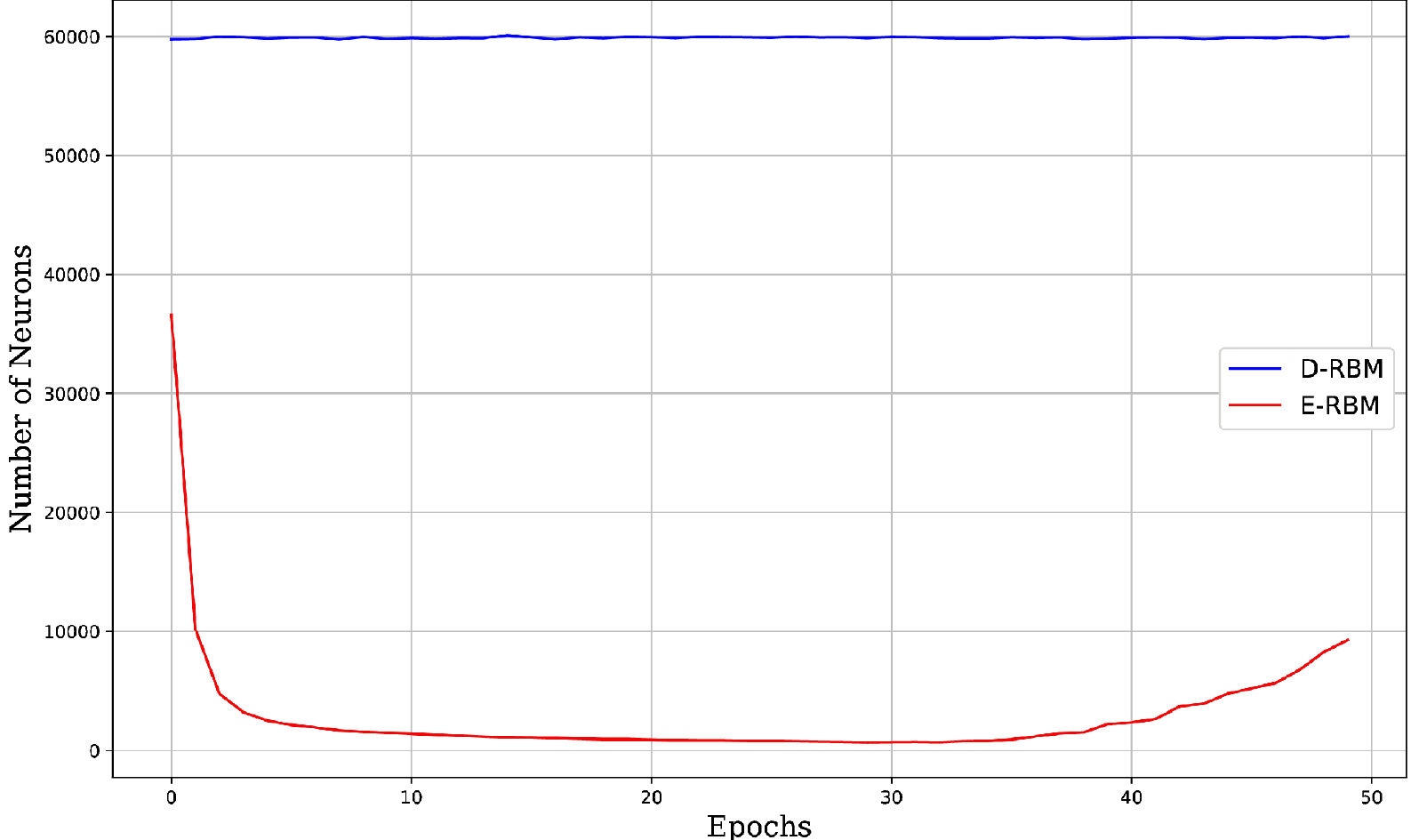}
\caption{Mean number of neurons dropped over Kuzushiji-MNIST's training set.}
\label{f.kmnist_off}
\end{figure}

The third architecture ($M_a$) is used here to illustrate how the E-Dropout affects the number of neurons turned off in the training process for the three datasets, as shown in Figures~\ref{f.mnist_off},~\ref{f.fash_off} and~\ref{f.kmnist_off}. For clarity, D-RBM tends to turn off $60,000$ ($n*p*60,000\_images/mini-batch$) neurons on every epoch, while E-RBM considers the neurons activation and the system energy batch-by-batch, and therefore, does not have a ``mean value''.

These results indicate that E-Dropout behavior depends on the dataset since it considers the relationship between neurons activation and the system's energy derived from the data itself.


Considering the MNIST dataset, the E-Dropout starts by almost turning off the same amount of neurons that the standard Dropout, and slowly decrease this value over the epochs.
For the Fashion-MNIST dataset, the E-Dropout starts with almost all neurons and starts dropping them out similar to a sigmoid function shape. On the other hand, considering the Kuzushiji-MNIST dataset, the E-Dropout starts by turning off approximately $37,000$ neurons and rapidly decreasing these values, while increasing the number of dropped out neurons in the last epochs.

\begin{figure}[!ht]
    \centering
    \begin{tabular}{ccc}
    \includegraphics[scale=1.1]{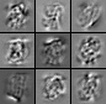} &
    \includegraphics[scale=1.1]{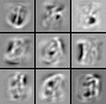} &
    \includegraphics[scale=1.1]{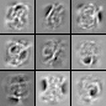} \\
    (I-A) & (II-A) & (III-A) \\
    \end{tabular}    
    \begin{tabular}{ccc}
    \includegraphics[scale=1.1]{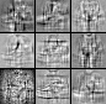} &
	\includegraphics[scale=1.1]{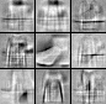} &
	\includegraphics[scale=1.1]{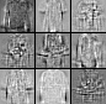} \\    
    (I-B) & (II-B) & (III-B) \\
    \end{tabular}
    \begin{tabular}{ccc}
    \includegraphics[scale=1.1]{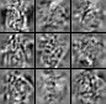} &
	\includegraphics[scale=1.1]{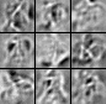} &
	\includegraphics[scale=1.1]{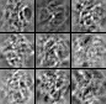} \\
	(I-C) & (II-C) & (III-C)    
    \end{tabular}
    \caption{$M_c$ - MNIST subset of learned weights: (I-A) RBM, (II-A) D-RBM and (III-A) E-RBM. Fashion-MNIST subset of learned weights: (I-B) RBM, (II-B) D-RBM and (III-B) E-RBM.
    Kuzushiji-MNIST subset of learned weights: (I-C) RBM, (II-C) D-RBM and (III-C) E-RBM.}
    \label{f.w8_mnist}
\end{figure}

Regarding the weights learned by the models, Figure~\ref{f.w8_mnist} depict a subset of model $M_c$ for MNIST (I-A, II-A, III-A), Fashion-MNIST (I-B, II-B, III-B), and Kuzushiji-MNIST (I-C, II-C, III-C) datasets, respectively. Overall, the D-RBM provides some sparsity but less ``clear'' weights, representing any images' details. The RBM provides a fair representation of these details, and even though, in some cases, it is clear that its weights are less informative. Finally, the E-RBM portrays more accurate images representation, mainly the high-frequency ones, such as the inner drawings. 

Additionally, one can establish a parallel with the temperature regularization effect showed by \cite{Passos:18} and \cite{Li:16}, in which low temperatures forces de connections to small values, providing network sparsity at the step that improves the lower bound in the learning process. Such behavior is interesting since the E-RBM was encouraged to prevent co-adaptations selectively.

\begin{figure}[!ht]
    \centering
    \begin{tabular}{cccc}
    \includegraphics[scale=1]{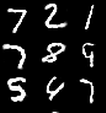} &
    \includegraphics[scale=1.05]{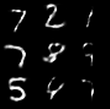} &
    \includegraphics[scale=1.01]{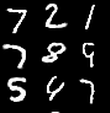} &
    \includegraphics[scale=1.05]{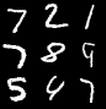} \\
    (I) & (II) & (III) & (IV)
    \end{tabular}
    \caption{$M_c$ - MNIST subset of reconstructed images: (I) RBM, (II) D-RBM, (III) E-RBM and (IV) Original.}
    \label{f.rec_mnist}
\end{figure}

\begin{figure}[!ht]
    \centering
    \begin{tabular}{cccc}
    \includegraphics[scale=1]{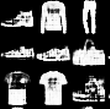} &
    \includegraphics[scale=1]{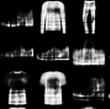} &
    \includegraphics[scale=1]{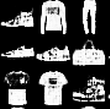} &
    \includegraphics[scale=1]{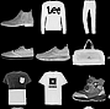} \\
    (I) & (II) & (III) & (IV)
    \end{tabular}
    \caption{$M_c$ - Fashion-MNIST subset of reconstructed images: (I) RBM, (II) D-RBM, (III) E-RBM and (IV) Original.}
    \label{f.rec_fash}
\end{figure}

\begin{figure}[!ht]
    \centering
    \begin{tabular}{cccc}
    \includegraphics[scale=1]{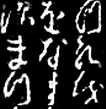} &
    \includegraphics[scale=1]{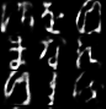} &
    \includegraphics[scale=1]{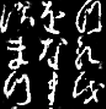} &
    \includegraphics[scale=1]{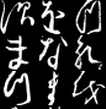} \\
    (I) & (II) & (III) & (IV)
    \end{tabular}
    \caption{$M_c$ - Kuzushiji-MNIST subset of reconstructed images: (I) RBM, (II) D-RBM, (III) E-RBM and (IV) Original.}
    \label{f.rec_kmnist}
\end{figure}

Furthermore, Figures~\ref{f.rec_mnist},~\ref{f.rec_fash} and~\ref{f.rec_kmnist} depict a subset of model $M_c$ reconstructed images over MNIST, Fashion-MNIST, and Kuzushiji-MNIST datasets, respectively. 

Finally, Table \ref{t.mean_time} shows the computational burden over all methods and architectures, regarding $50$ epoch of training. It is essential to highlight that all datasets consume almost the same computational load due to the same characteristics, and, for that, it was summarized in one general table. The mean and standard deviation are from the ten repetitions taken from the experiments.

\begin{table}[!ht]
	\renewcommand{\arraystretch}{1.75}
	\caption{Mean time in seconds and their respective standard deviation.}
	\label{t.mean_time}
	\centering
	\begin{tabular}{lcccc}
		\toprule
		\textbf{Technique} & $\mathbf{M_a}$ & $\mathbf{M_b}$ & $\mathbf{M_c}$ & $\mathbf{M_d}$\\ \midrule
		RBM~\cite{Hinton:02} & 250 $\pm$ 3 & 250 $\pm$ 3 & 250 $\pm$ 3 & 250 $\pm$ 3 \\
		W-RBM~\cite{Hinton:12} & 250 $\pm$ 3 & 250 $\pm$ 3 & 250 $\pm$ 3 & 250 $\pm$ 3 \\
		D-RBM~\cite{Srivastava:14} & 250 $\pm$ 3 & 250 $\pm$ 3 & 250 $\pm$ 3 & 250 $\pm$ 3 \\
		DC-RBM~\cite{wan2013} & 1050 $\pm$ 3 & 1100 $\pm$ 3 & 1100 $\pm$ 3 & 1100 $\pm$ 3 \\
		E-RBM & 300 $\pm$ 3 & 300 $\pm$ 4 & 300 $\pm$ 4 & 300 $\pm$ 4 \\
		\bottomrule
	\end{tabular}
\end{table}

Table~\ref{t.mean_time} shows that E-RBM has a little more computational load than RBM, W-RBM, and D-RBM, but considering a high number of training epochs. On the other hand, the DC-RBM was the more power-consume model since the DropConnect needs to sample a weight mask for every instance on the mini-batch. In summary, the improvement achieved by the E-RBM in the image reconstruction task depicted in previous sections overcome the slightly worst performance in processing time against the simpler baselines.
\section{Conclusion}
\label{s.conclusion}

This article proposed a new regularization method, known as energy-based Dropout, an enhanced parameterless version of the traditional Dropout. Based on physical principles, it creates a direct correlation between the system's energy and its hidden neurons, denoted as Importance Level ($\cal{I}$). Furthermore, as Restricted Boltzmann Machines are also physical-based neural networks, they were considered the perfect architecture to validate the proposed approach.

The energy-based Dropout was validated in Restricted Boltzmann Machines through a binary image reconstruction task. Three well-known literature, datasets, MNIST, Fashion-MNIST, and Kuzushiji-MNIST, were employed to validate the proposed approach. Considering the experimental results discussed in the paper, one can observe that the energy-based Dropout proved to be a suitable regularization technique, obtaining significantly better SSIM rates than its counterpart Dropout in all three datasets. Additionally, when comparing the energy-based Dropout to the standard RBM, it outperformed the latter in two out of three datasets, being slightly worse in the one that it could not achieve the best result. Moreover, it is possible to perceive that the weights learned by the energy-Dropout approach were able to recognize different patterns and high-frequency details, besides had less sharp edges when compared to the standard RBM and Dropout-based RBM.

When comparing all the employed techniques, more demanding tasks benefit more from the energy-based Dropout than easier ones, i.e., tasks with higher reconstruction errors seem to achieved the best result when using the energy-based Dropout. Moreover, when comparing the proposed method and the standard one, the proposed regularization obtained significantly better results, reinforcing its capacity to improve RBMs' learning procedure.

Regarding future works, we aim at expanding some concepts of the energy-Dropout regularization technique to the classification task and other suitable machine learning algorithms, such as Deep Belief Networks (DBNs) and Deep Boltzmann Machines (DBMs).


\section*{Acknowledgements}
The authors are grateful to S\~ao Paulo Research Foundation (FAPESP) grants \#14/12236-1, \#2019/07825-1 and \#2019/02205-5. Also, VHCA received support from the Brazilian National Council for Research and Development (CNPq), grants \#304315/2017-6 and \#430274/2018-1. JPP also acknowledges CNPq grants \#307066/2017-7 and \#427968/2018-6.

\ifCLASSOPTIONcaptionsoff
  \newpage
\fi

\bibliographystyle{IEEEtran}
\bibliography{references}

\begin{IEEEbiography}[{\includegraphics[width=1in,height=1.25in,clip,keepaspectratio]{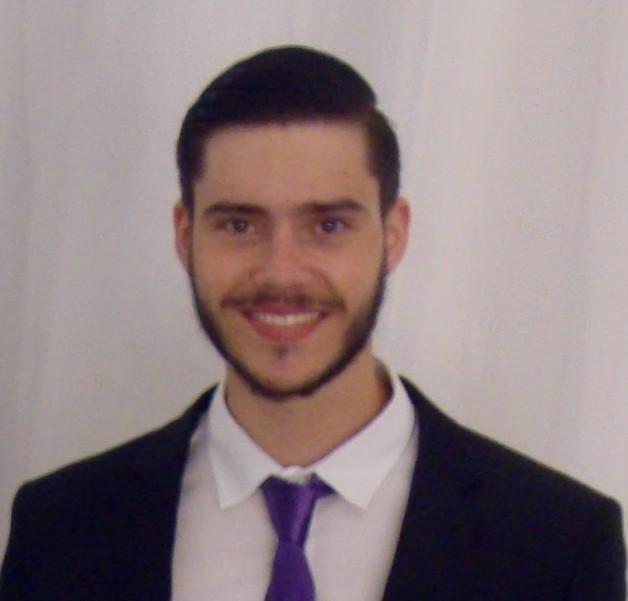}}]{Mateus Roder} Mateus Roder is a Bachelor in Manufacturing Engineering at S\~ao Paulo State University (UNESP), Itapeva-SP (2018), and a former Scientific Initiation FAPESP's scholarship holder, focusing on image processing and classification, machine learning algorithms, and meta-heuristic optimization. Currently, is a student and FAPESP's scholarship holder in Master of Computer Science at S\~ao Paulo State University, FC/Bauru, focusing on restricted Boltzmann machines and deep learning. Email: mateus.roder@unesp.br
\end{IEEEbiography}

\begin{IEEEbiography}[{\includegraphics[width=1in,height=1.25in,clip,keepaspectratio]{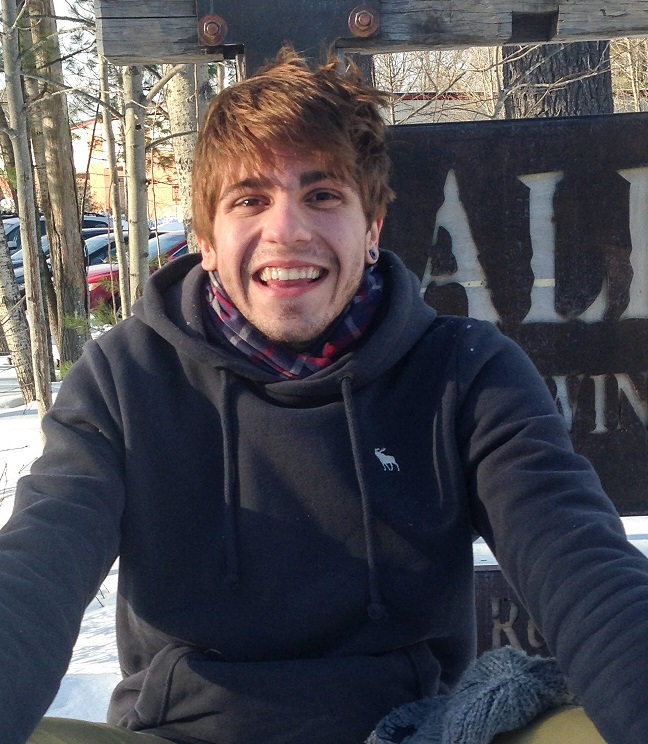}}]{Gustavo Henrique de Rosa} is a Bachelor in Computer Science at S\~ao Paulo State University (UNESP), FC/Bauru (2016), and a former Scientific Initiation FAPESP's scholarship holder with an internship at the Harvard University, focusing on image processing formulations, pattern recognition, pattern classification, machine learning algorithms, and meta-heuristic optimization. Master of Science in Computer Science at S\~ao Paulo State University, IBILCE/Rio Preto (2018), and a former Master of Science FAPESP's scholarship holder with an internship at the University of Virginia, focusing on deep learning and meta-heuristic optimization. Currently, as a Ph.D. student in Computer Science at S\~ao Paulo State University, FC/Bauru, and a Ph.D. FAPESP's scholarship holder, focusing on natural language processing and adversarial learning. Email: gustavo.rosa@unesp.br 
\end{IEEEbiography}

\begin{IEEEbiography}[{\includegraphics[width=1in,height=1.25in,clip,keepaspectratio]{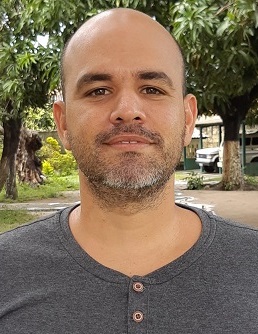}}] {Victor Hugo C. de Albuquerque} [M’17, SM’19] is a professor and senior researcher at the ARMTEC Tecnologia em Robótica, Brazil. He has a Ph.D in Mechanical Engineering from the Federal University of Paraíba (UFPB, 2010), an MSc in Teleinformatics Engineering from the Federal University of Ceará (UFC, 2007), and he graduated in Mechatronics Engineering at the Federal Center of Technological Education of Ceará (CEFETCE, 2006). He is a specialist, mainly, in IoT, Machine/Deep Learning, Pattern Recognition, Robotic.
\end{IEEEbiography}

\begin{IEEEbiography}[{\includegraphics[width=1in,height=1.25in,clip,keepaspectratio]{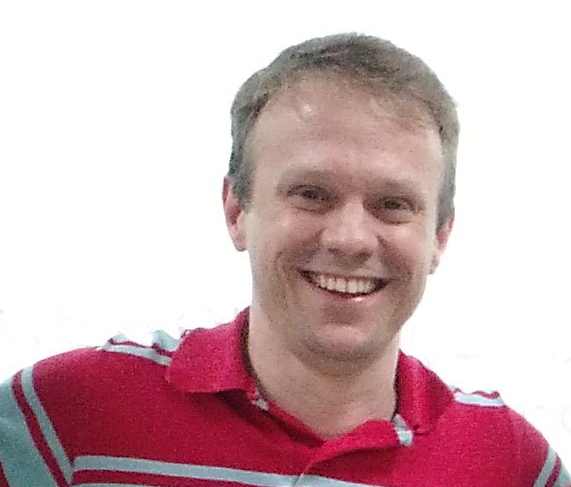}}]{Andr\'e Rossi} André L.D. Rossi received his B.Sc. degree in Computer Science from the Universidade Estadual de Londrina (UEL), Brazil, and his M.Sc. and Ph.D. degrees in Computer Science from  University of São Paulo (USP), Brazil. André Rossi is currently an Associate Professor at the São Paulo State University (UNESP), Brazil. His main research interests are Machine Learning and Computer Vision. Email: andre.rossi@unesp.br 
\end{IEEEbiography}

\begin{IEEEbiography}[{\includegraphics[width=1in,height=1.25in,clip,keepaspectratio]{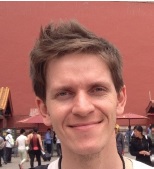}}]{Jo\~ao P. Papa} [SM'17] received his B.Sc.in Information Systems from the S\~{a}o Paulo State University (UNESP), SP, Brazil. In 2005, he received his M.Sc. in Computer Science from the Federal University of S\~ao Carlos, SP, Brazil. In 2008, he received his Ph.D. in Computer Science from the University of Campinas, SP, Brazil. During 2008-2009, he had worked as a post-doctorate researcher at the same institute, and during 2014-2015 he worked as a visiting scholar at Harvard University. He has been a Professor at the Computer Science Department, S\~ao Paulo, State University, since 2009. He was also the recipient of the Alexander von Humboldt research fellowship in 2017. Email: joao.papa@unesp.br 
\end{IEEEbiography}

\end{document}